\documentclass[10pt,twocolumn,letterpaper]{article}

\usepackage{wacv}
\usepackage{times}
\usepackage{epsfig}
\usepackage{graphicx}
\usepackage{amsmath}
\usepackage{amssymb}
\usepackage[accsupp]{axessibility}  

\usepackage{fdsymbol}
\usepackage{algorithm}
\usepackage{algorithmic}
\usepackage{bm, multirow}
\usepackage{enumitem}

\usepackage{color,soul}
\usepackage{float}

\usepackage{tabularx} 
\usepackage{rotating}
\newcolumntype{L}[1]{>{\hsize=#1\hsize\raggedright\arraybackslash}X}%
\newcolumntype{R}[1]{>{\hsize=#1\hsize\raggedleft\arraybackslash}X}%
\newcolumntype{C}[1]{>{\hsize=#1\hsize\centering\arraybackslash}X}%

\newcommand{\bx}{\bm{x}}
\newcommand{\by}{\bm{y}}
\newcommand{\bz}{\bm{z}}

\graphicspath{{figs/}}
\usepackage[pagebackref=true,breaklinks=true,colorlinks,bookmarks=false]{hyperref}

\def\wacvPaperID{1001} 

\wacvfinalcopy 

\ifwacvfinal
\fi

\ifwacvfinal
\usepackage[breaklinks=true,bookmarks=false]{hyperref}
\else
\usepackage[pagebackref=true,breaklinks=true,colorlinks,bookmarks=false]{hyperref}
\fi

\begin{document}

\title{Occlusion-Robust Object Pose Estimation with Holistic Representation}

\author{Bo Chen\\
The University of Adelaide\\
{\tt\small bo.chen@adelaide.edu.au}
\and
Tat-Jun Chin\\
The University of Adelaide\\
{\tt\small tat-jun.chin@adelaide.edu.au}

\and
Marius Klimavicius\\
Blackswan Technologies\\
{\tt\small marius@blackswan.ltd}

}

\maketitle
\thispagestyle{empty}

\begin{abstract}
Practical object pose estimation demands robustness against occlusions to the target object. State-of-the-art (SOTA) object pose estimators take a two-stage approach, where the first stage predicts 2D landmarks using a deep network and the second stage solves for 6DOF pose from 2D-3D correspondences. Albeit widely adopted, such two-stage approaches could suffer from novel occlusions when generalising and weak landmark coherence due to disrupted features. To address these issues, we develop a novel occlude-and-blackout batch augmentation technique to learn occlusion-robust deep features, and a multi-precision supervision architecture to encourage holistic pose representation learning for accurate and coherent landmark predictions. We perform careful ablation tests to verify the impact of our innovations and compare our method to SOTA pose estimators. Without the need of any post-processing or refinement, our method exhibits superior performance on the LINEMOD dataset. On the YCB-Video dataset our method outperforms all non-refinement methods in terms of the ADD(-S) metric. We also demonstrate the high data-efficiency of our method. Our code is available at \url{http://github.com/BoChenYS/ROPE}
\end{abstract}

\section{Introduction}

\begin{figure}
    \centering
    \includegraphics[width=\linewidth]{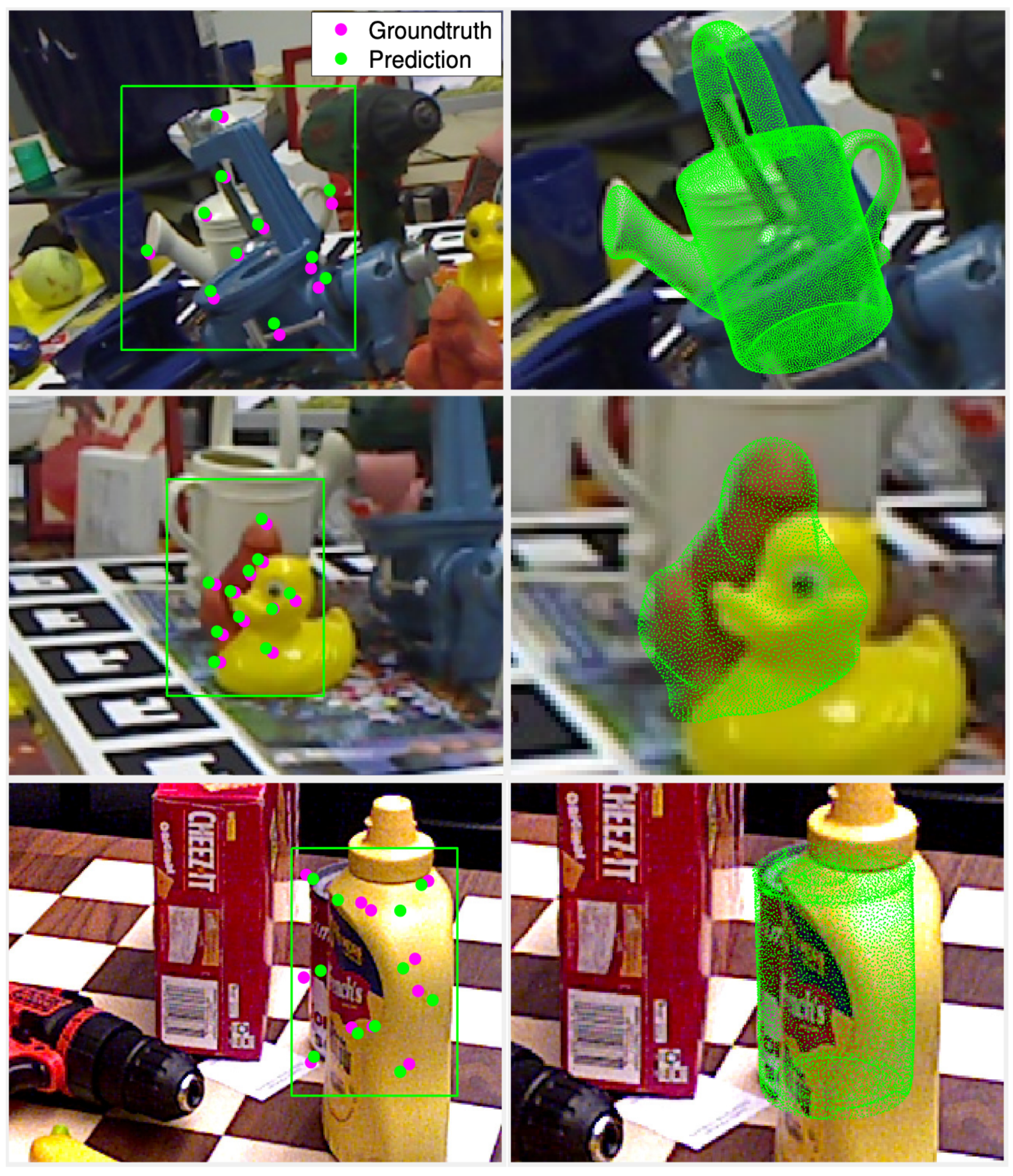}
    \caption{Qualitative results of our robust object pose estimation on the Occluded-LINEMOD (top and middle) and the YCB-Video (bottom) datasets.  \textbf{Left}: prediction of bounding boxes and landmarks of the target object in a test image (zoomed view). \textbf{Right}: prediction of 6DOF poses without post-processing or refinement.}
    \label{fig:punch}
\end{figure}

Object pose estimation is the task of inferring the relative orientation and position between the target object and the observer. Such inference is crucial in many vision applications such as robotic manipulation~\cite{Zuo2019craves, zhu2014single, collet2011moped}, augmented reality~\cite{marchand2015pose, crivellaro2018robust}, autonomous driving~\cite{chen2017multi, wu20196d, xu2018pointfusion} and spacecraft navigation~\cite{Cassinis2019review, sharma2018pose}. The problem can be simplified if depth information is available~\cite{michel2017global, wang2019densefusion, He2020pvn3d, chen2021fs}. However, depth sensors are not always practical. Pose estimation from images is thus an important research problem.

In this paper we consider the problem of object pose estimation from a single RGB image. Our focus lies in the base estimator, \ie, from input image to the output pose, before any refinement step. For the base estimator, a number of works~\cite{Kehl2017ssd,Xiang2018posecnn,poirson2016fast,do2018deep} adopt direct regression approaches which map the input image that contains the target object to its 6 DOF pose. However, such approaches tend to be sensitive to occlusions and are observed to be similar to performing image retrieval~\cite{sattler2019understanding}.



Rather than directly regressing the pose, two-stage approaches~\cite{Hu2019segmentation, Jafari2018ipose, Li2019cdpn, Oberweger2018making, Park2019pix2pose, Peng2019pvnet, rad2017bb8, Zakharov2019dpod, song2020hybridpose, pavlakos20176, tekin2018real} first predict landmarks on the object to establish 2D-3D correspondences, then use a Perspective-n-Point (PnP) like algorithm to solve for the pose. Previous results suggest that two-stage methods are generally more accurate~\cite{Oberweger2018making, Hu2020single}. Their strengths derive from training the model with richer supervision signals (\ie, groundtruth landmarks) rather than just the pose, and injecting tolerance towards inaccurate landmark predictions by robust PnP.

However, two-stage approaches are not intrinsically immune to occlusion. Current works to improve robustness often take the pixel-wise or patch-wise approach~\cite{Peng2019pvnet, Oberweger2018making, Hu2019segmentation, Jafari2018ipose, Li2019cdpn, Park2019pix2pose}, \ie, generating an ensemble of predictions from each image pixel or patch, and aggregate them to obtain a more robust final prediction. Although ensembling can mitigate some occlusion-induced inaccuracies, landmark coherence are easily disrupted by large and novel occlusions, because the network predicts landmarks independently and consistency is only imposed by the PnP algorithm, which is not part of the network~\cite{Hu2020single}. 

In this paper, we aim to address the shortcomings of current two-stage approaches. Firstly, we enforce occlusion-robust feature learning to enable models to deal with novel and severe occlusions. Secondly, a good pose representation should produce landmarks that are consistent to the object shape, rather than predicting individual landmarks independently. To this end, during model training we encourage a holistic pose representation learning in order to strengthen the connections between landmark predictions and enhance their coherence.

\paragraph{Our contributions} We propose the Robust Object Pose Estimation (ROPE) framework which 
achieves excellent robustness against occlusions without the need of pose refinement. 
As shown in Figure~\ref{fig:punch}, our model predicts landmarks and pose robustly without any post-processing.  


To enforce occlusion-robust feature learning, we combine hide-and-seek~\cite{Singh2017hide}, random erasing~\cite{Zhong2020random} and batch augmentation~\cite{Hoffer2020augment} and propose a occlude-and-blackout batch augmentation technique for model training. 
To encourage the model to learn holistic pose representations, we propose a multi-precision supervision architecture, which boosts the model's ability to extrapolate occluded object parts, leading to spatially more accurate and structurally more coherent landmark predictions. 
To alleviate the need for pose refinement we further utilise the multi-precision supervision architecture to filter landmark predictions with a simple verification step. 

We conduct extensive experiments to verify the efficacy of the proposed techniques, and compare our method to SOTA object pose estimators. In terms of the ADD(-S) metirc, our method outperforms all contestants on LINEMOD~\cite{Hinterstoisser2012model} and all non-refinement methods on YCB-Video~\cite{Xiang2018posecnn}. Without any refinement, it is also competitive to SOTA methods that includes a refinement step. Compared to methods that relies on large amount of synthetic training images, we show that ROPE is highly data-efficient. 





\section{Related works}

Traditional object pose estimation methods~\cite{gu2010discriminative, hinterstoisser2011gradient, huttenlocher1993comparing, Hinterstoisser2012model, lepetit2005monocular, lowe1999object} rely on hand-crafted features or template matching techniques, which are susceptible to occlusions or other appearance change. Recent advancements of deep learning has nurtured a lot of learning-based methods. We briefly survey a few prominent works from one-stage, two-stage and other methods. 

PoseNet~\cite{kendall2015posenet} was a pioneer work on using a deep model to directly regress the 6DOF from an image. Although it was proposed for camera localisation rather than object pose estimation, its principle applies to both tasks. SSD-6D~\cite{Kehl2017ssd} combines an SSD detector~\cite{Liu2016ssd} and a pose regressor in a single network. RenderForCNN~\cite{su2015render} uses an image renderer to synthesize training images as well as groundtruth pose for training a pose regressor. 

Compared to one-stage approaches, two-stage methods typically predicts intermediate features in the first stage, and then solve for the pose in the second stage. This mechanism receives more attention because its intermediate feature learning facilitates more potential improvements. For example, Tekin \etal~\cite{tekin2018real} apply the YOLO object detector~\cite{redmon2017yolo9000} in the first stage to predict object landmarks. 
Hu \etal~\cite{Hu2019segmentation} predict landmark locations for each small patch of the input image. They then aggregate all patch predictions to establish 2D-3D correspondences for solving the pose. Oberweger \etal~\cite{Oberweger2018making} on the other hand, only use patches of images to train the landmark predictor. The idea is that at least some patches are not corrupted by the occluder and they could produce accurate landmark heatmaps. The ensemble of heatmaps predicted from many patches are combined to obtain final landmarks. PVNet~\cite{Peng2019pvnet} predicts the object mask and, for each pixel within the mask, unit vectors that points to the landmarks. It then utilises a generalised Hough voting scheme~\cite{ballard1981generalizing} to determine the distribution of the landmarks. 


There are other notable works tackling object pose estimation from different perspectives. Sundermeyer \etal~\cite{sundermeyer2020augmented, sundermeyer2020multi} use autoencoders to learn implicit pose representations by reconstructing the input objects. Cai and Reid~\cite{Cai2020reconstruct} propose a 3D model-free pose estimator via 2D-3D mapping. To make two-stage methods into a single stage pipeline, Hu \etal~\cite{Hu2020single} and Wang \etal~\cite{wang2021gdr} propose deep architectures to replace the PnP algorithm in the second stage, while Chen \etal~\cite{Chen2020end} propose a differentiable PnP method to achieve end-to-end learning.

\begin{figure*}[t]
    \centering
    \includegraphics[width=0.9\linewidth]{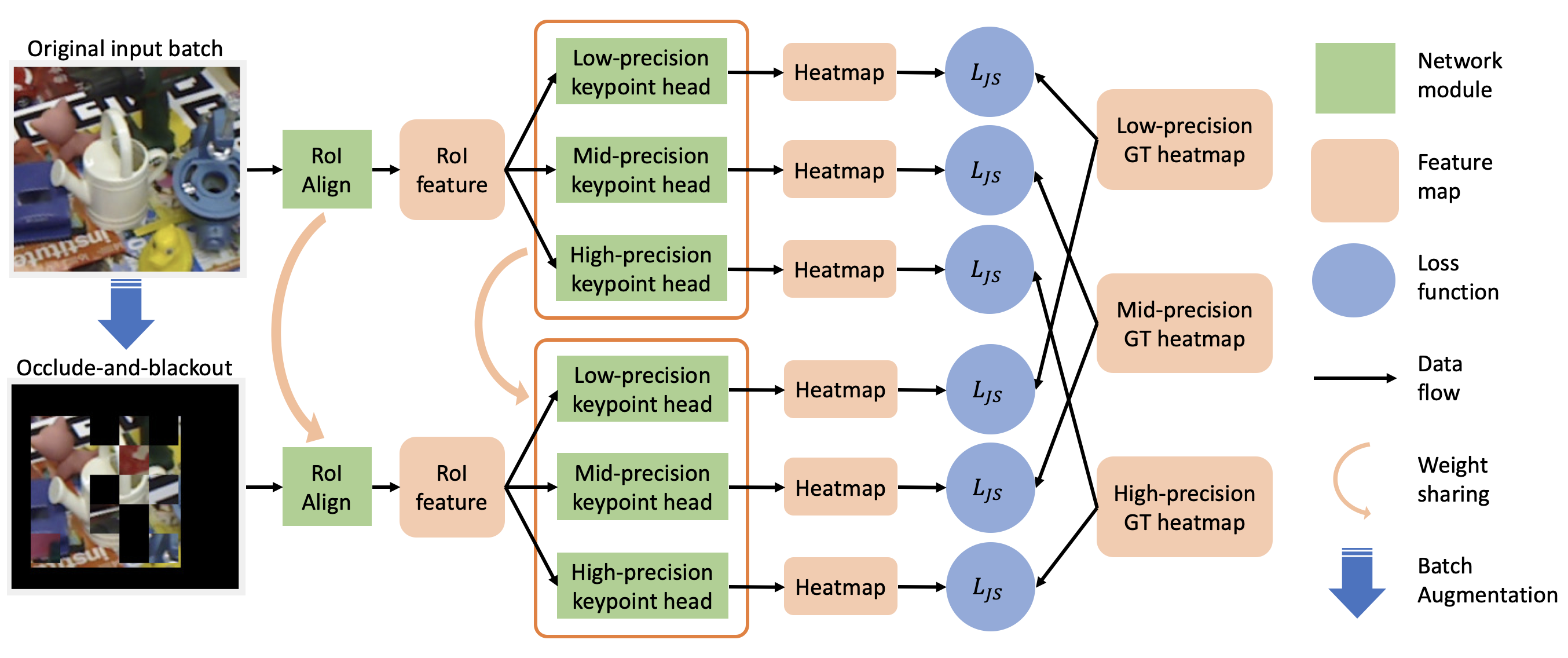}
    \caption{Illustration of an occlude-and-blackout augmented example and the architecture of our heatmap prediction network. For clarity, the backbone and the RPN are represented in the RoI Align module, other modules in the Mask R-CNN framework such as the box head, as well as relevant losses, are not shown. Our model replaces the original mask head with three keypoint heads.}
    \label{fig:hm_arch}
\end{figure*}

\section{The ROPE framework}

We focus on the problem of 6DOF object pose estimation from a single RGB image. Given an image $I$ and a known 3D point cloud $\{ \bz_i \}^{n}_{i=1}$ of the target object, we first predict a set of 2D landmarks $\{ \bx_i \}^{n}_{i=1}$ in $I$ that correspond to the point cloud, then solve the pose $\by$ via a RANSAC-based PnP solver from filtered 2D-3D correspondences. 



\subsection{Robust landmark prediction}

Our 2D landmark prediction is based on the Mask R-CNN~\cite{He2018mask} framework. The specific architecture and training scheme are shown in Figure~\ref{fig:hm_arch}. A basic improvement is substituting the original backbone network with HRNet~\cite{Sun2019deep, Wang2020deep} to exploit its high-resolution feature maps which preserve rich semantic information and increase spatial accuracy. Next, we describe two key innovations to boost occlusion robustness and landmark coherence.

\subsubsection{Occlude-and-blackout batch augmentation}

Fundamentally, pose estimation for the typical 3D object will suffer from the problem of self-occlusion. Landmarks that are at the opposite side of the object would be hard to predict since their visual features are hidden. In fact, a practical pose estimator must also contend with additional occlusions due to, \eg, other objects or scene elements that further conceal part of the target object from view. It is thus important that the landmark predictor infers the robust pose information from potentially different kinds occlusions imposed on the object.


Inspired by the ideas of random erasing~\cite{Zhong2020random}, hide-and-seek~\cite{Singh2017hide}, and batch augmentation~\cite{Hoffer2020augment} (all not originally developed for pose estimation), we develop a novel Occlude-and-blackout Batch Augmentation (OBA) to promote robust landmark prediction under occlusion. For each training batch, after performing regular data augmentations including rotation, translation, scaling and color jitter, we extend the batch by including a copy of itself with extra augmentations, namely, occlude and blackout. Similar to hide-and-seek, we divide the image region enveloped by the object bounding box into a grid of patches and replace each patch, under certain probability, with either noise or a random patch elsewhere from the same image. We then blackout everything outside of the object bounding box. An example is shown in Figure~\ref{fig:hm_arch}. 

With random occlusions the network is forced to infer the pose information from a partial view of the object. Erasing the background helps reducing overfitting and enhance generalisability. Moreover, the OBA augmented images are fed to the network with the original ones in the same batch, and supervised by the same groundtruth labels. This encourages the network to learn occlusion-robust and background-invariant representations.

If the potential occluders are known beforehand, injecting  occluder specific information in the training phase can significantly improve performance~\cite{Oberweger2018making}. However this knowledge is often not available in practice. Compared to methods that augment training images with known objects~\cite{Jafari2018ipose, li2017deep, alhaija2017augmented}, our method is occluder-agnostic yet it generalises well in the testing sets.

\subsubsection{Multi-precision supervision}

Current heatmap-based landmark prediction networks use a single groundtruth Gaussian heatmap per landmark for training. The variance of these heatmaps is a hyper parameter which requires careful tuning: a smaller variance may increase prediction accuracy for each individual landmark however risk structural inconsistency in the case of occlusion, due to the lack of holistic understanding of the object pose. To address this issue we propose a Multi-Precision Supervision (MPS) architecture: using three keypoint heads to predict groundtruth Gaussian heatmaps with different variance. 

\begin{figure}[t]
    \centering
    \includegraphics[width=\linewidth]{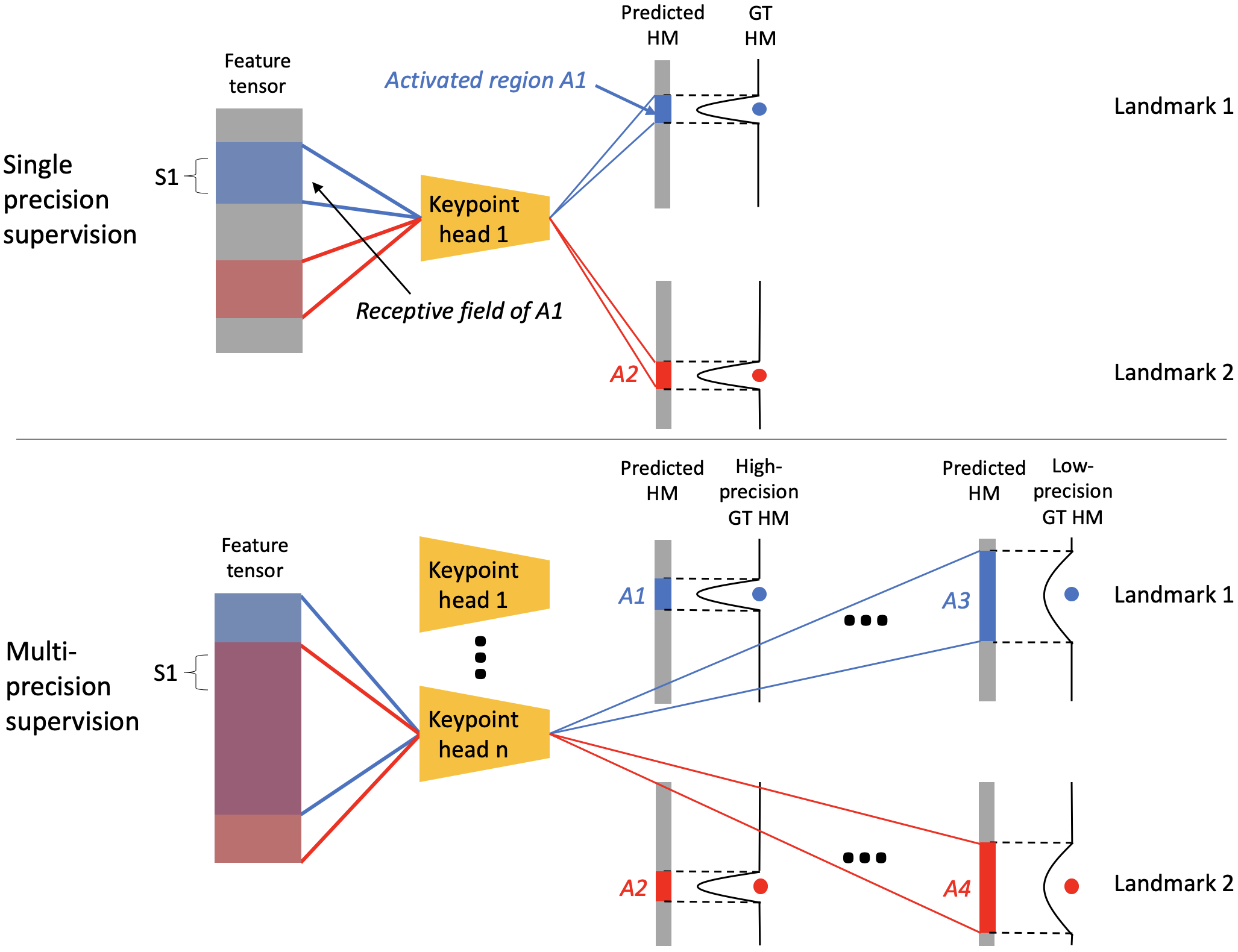}
    \caption{Conceptual illustration of holistic representation learning via MPS. Note the difference on the information learned by the feature section $S1$.}
    \label{fig:sps_vs_mps}
\end{figure}

In Mask R-CNN, the output feature map of the backbone is aligned with RoI proposals and the RoI features are then passed to the mask head. We replace the mask head with three keypoint heads to regress the landmark heatmaps, as shown in Figure~\ref{fig:hm_arch}. Each keypoint head consists of 8 convolutional layers and 2 upsampling layers. 

In the training phase, the groundtruth heatmaps $\Phi^*$ are constructed as 2D Gaussian feature maps centred on groundtruth 2D landmarks $\bx^*$ and spreading with variance $\sigma^2$. We use $\sigma$ equal to 8, 3 and 1.5 pixels respectively for the three keypoint heads, thus creating low, medium and high precision target heatmaps $\Phi^*$. The loss function is
\begin{equation}
    L_{JS} = \text{JSD}(\phi(\Phi), \Phi^*),
\end{equation}
where JSD$(\cdot)$ is the Jensen–Shannon divergence~\cite{Fuglede2004jensen} and $\phi(\cdot)$ is the channel-wise softmax function, \ie, each channel is normalised to be a probability distribution over the pixels. 

In the testing phase, we only use the predicted heatmaps $\Phi$ from the high-precision keypoint head to obtain the landmark coordinates $\bx$. Instead of simply taking the ``argmax'' of $\Phi$ as $\bx$, we treat the normalised heatmaps $\phi(\Phi)$ as probability maps and take their spatial expectations as $\bx$. This has two advantages over the ``argmax'' approach: it has higher accuracy because it is continuous rather than discrete; it is more robust to outlying pixel values. 

Although only the high-precision heatmaps are used to compute the landmark coordinates, the medium and low-precision keypoint heads play an important role in the pipeline. Firstly, having target heatmaps with different variances $\sigma^2$ helps the model adapt to objects of different sizes. This also relieves the need for tuning $\sigma$ as a hyper parameter for each object. Secondly, heatmaps from the medium-precision keypoint head are used as an auxiliary for filtering predicted landmarks, as will be explained in the next subsection. Lastly and most importantly, MPS boosts holistic representation learning in the feature maps and increases landmark coherence. An conceptual illustration is shown in Figure~\ref{fig:sps_vs_mps}. 

In Figure~\ref{fig:sps_vs_mps}, we take one section of the feature tensor $S1$ for examination. With single precision supervision, $S1$ is only responsible for activating the region $A1$ in the predicted heatmap of Landmark 1. It does not learn useful information about Landmark 2. In the MPS scenario, besides learning about Landmark 1 via $A1$ and $A3$, $S1$ is also exposed to the receptive field of $A4$ from Landmark 2. This enforces $S1$ to incorporate relevant information and become more ``aware'' of the location of Landmark 2. The overall effect is that, each part of the feature tensor not only learns the necessary information to predict a local landmark, but also integrates knowledge of other landmarks to understand a wider context, thus learns a more holistic representation of the target object pose. 

\begin{figure}[t]
    \centering
    \includegraphics[width=\linewidth]{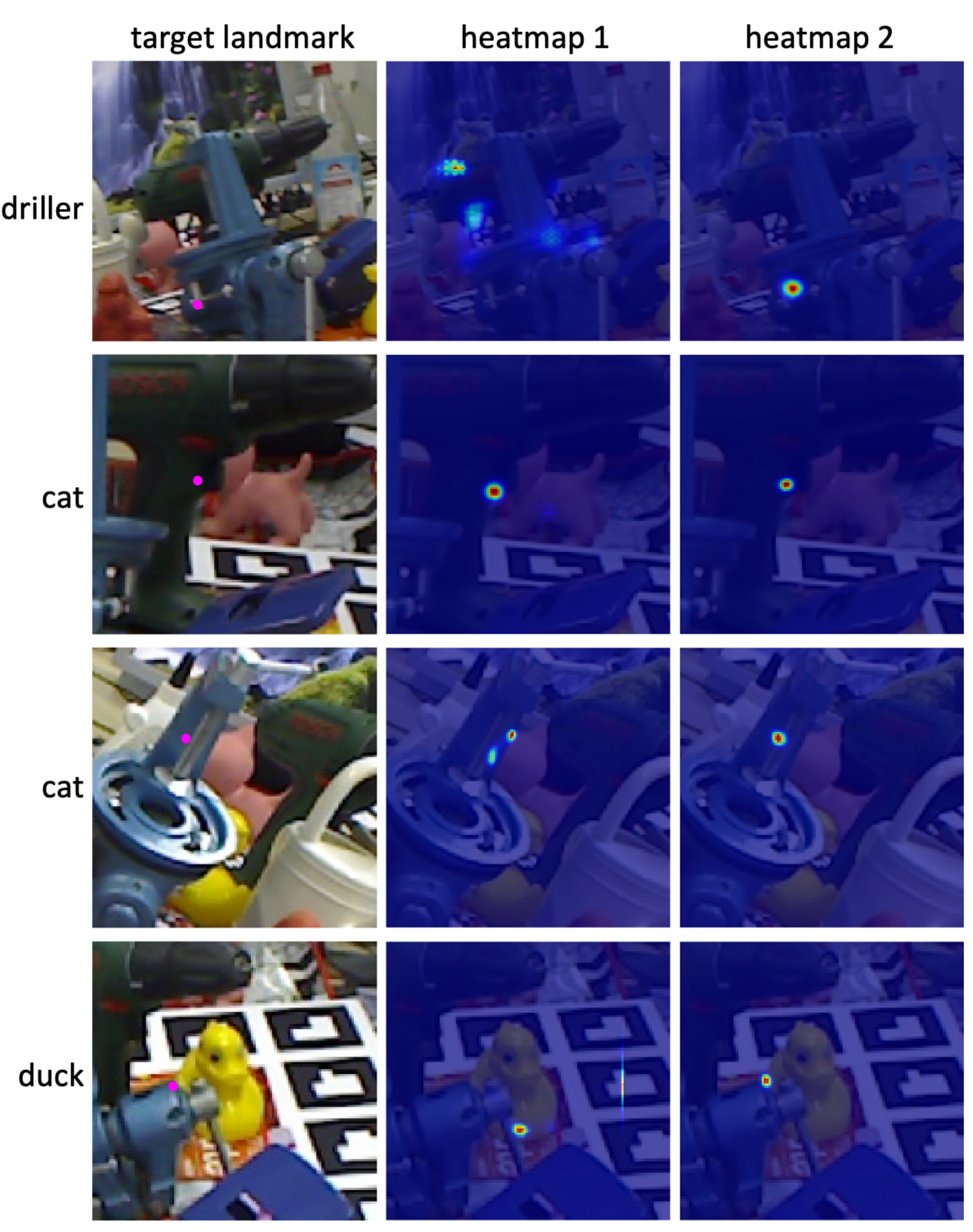}
    \caption{The effect of holistic representation learning in heatmap prediction. Predictions of heatmap 1 are from a model (MV1) trained without MPS while those of heatmap 2 are from the full model (original) with MPS. Details of models (MV1 and original) are provided in Section~\ref{sec:mvs}. }
    \label{fig:hm_comp}
\end{figure}

A holistic representation enables heatmap predictions to be more robust against occlusions. As shown in Figure~\ref{fig:hm_comp}, when trained without MPS, novel occlusions result in confused heatmap activations. On the other hand, a holistic representation learned via MPS is able to produces stable heatmaps for the occluded landmarks. This also boosts the structural consistency of landmark predictions as shown in Figure~\ref{fig:ablation1} and \ref{fig:mps}, which is further discussed in Section~\ref{sec:ablation}.


\subsection{Landmark filtering}
Many pose estimation pipelines include a refinement stage which is either optimisation-based \cite{Kehl2017ssd, Chen2019satellite, song2020hybridpose} or learning-based \cite{rad2017bb8, li2020deepim, labbe2020cosypose, Zakharov2019dpod}. While such post-processing is effective in boosting prediction accuracy, it adds additional computation burdens which is a disadvantage especially for real-time applications. In order to boost prediction accuracy while at the same time avoiding heavy post-processing computation, we make use of the multi-heads design of MPS for selecting high-quality landmark predictions before passing them to the PnP solver, thus alleviating the need for significant pose refinement. 

Specifically, for an image $I$, let $\{\bx_i\}$ denote the set of predicted landmark coordinates from the high-precision keypoint head, and $\{\bx_i^m\}$ denote the set of landmark coordinates predicted from the medium-precision keypoint head. We then select a subset
\begin{equation}
    \{\bx_i | \rVert \bx_i - \bx_i^m \rVert_2 \leq \epsilon \}
\end{equation}
for the PnP solver to compute the pose. In other words, a landmark prediction from the high-precision head will only be selected for the pose solver if it is verified by the corresponding medium-precision prediction, where $\epsilon$ is the verification threshold. In the case that the selected subset has fewer than 4 points, which is the minimum number required by a PnP solver, we then use the 4 points with the smallest $\rVert \bx_i - \bx_i^m \rVert_2$ values as the subset. 

While in this work we focus on the base pose estimator and report its performances without any refinement, our pipeline can be easily extended to stack one or multiple refiners such as \cite{Manhardt2018deep, Li2018deepim, sundermeyer2020multi}.

\section{Experiments}
In this section we conduct experiments to validate the effectiveness of ROPE as well as to compare it to SOTA methods of RGB image-based pose estimation.

\begin{figure*}[ht]
    \centering
    \includegraphics[width=\linewidth]{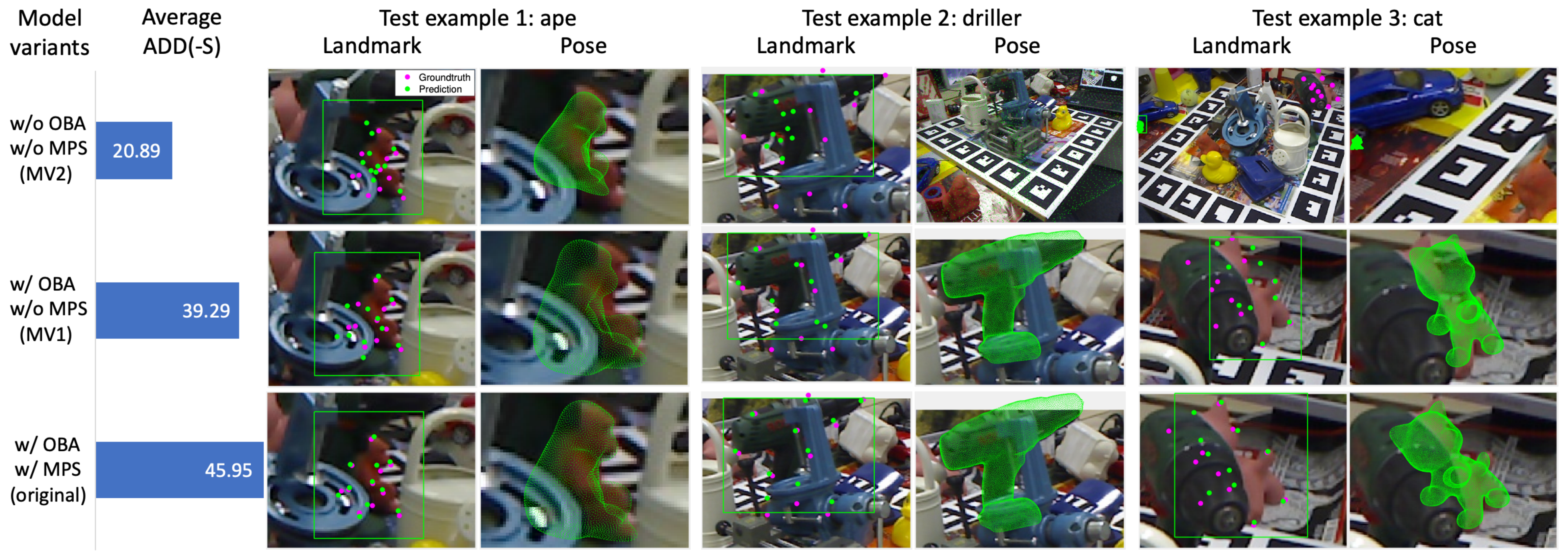}
    \caption{Comparing performances of model variants on the Occluded-LINEMOD dataset with qualitative examples. }
    \label{fig:ablation1}
\end{figure*}

\subsection{Datasets and metrics}
We choose the widely used LINEMOD~\cite{Hinterstoisser2012model}, its extension Occluded-LINEMOD~\cite{Brachmann2014learning} and the YCB-Video~\cite{Xiang2018posecnn} datasets for our experiments. 

For LINEMOD, we follow the convention of previous works~\cite{rad2017bb8, tekin2018real, Peng2019pvnet, Zakharov2019dpod} by using 15\% of the images of each object as training set and the remaining 85\% as testing set. The training images are selected in such a way that the relative rotation between them are larger than a threshold. For each object, we additionally use 1312 rendered images of the isolated object for training, which are obtained from~\cite{hodan2018bop}. For Occluded-LINEMOD the whole dataset is used for testing while images of the corresponding objects in LINEMOD, as well as the rendered images, are used for training. We also follow the protocol of \cite{Xiang2018posecnn, Oberweger2018making} for the YCB-Video dataset: we use 80 out of the 92 video sequences as well as the 80000 synthetic images for training, and test on 2949 key frames from the reserved 12 sequences.  


We report the ADD(-S) metric which combines the ADD metric~\cite{Hinterstoisser2012model} for asymmetric objects and the ADD-S metric~\cite{Xiang2018posecnn} for symmetric ones. The ADD metric computes the percentage of correctly estimated poses. A pose is considered correct if the object model points, when transformed by the predicted and groundtruth poses respectively, have an average distance of less 10\% of the model diameter. For ADD-S, this distance is instead computed based on the closest point distance. The ADD(-S) metric is preferred over the 2D projection metric~\cite{brachmann2016Uncertainty} because it directly measures the alignment discrepancy in 3D. 

For the YCB-Video dataset we also report the AUC metric proposed in~\cite{Xiang2018posecnn} and adopted in \cite{Oberweger2018making, Peng2019pvnet}. The AUC metric is the area under the ADD(-S) curve when varying the distance threshold for a pose to be deemed correct. We vary this threshold from 0 to 10 cm, in accordance with~\cite{Xiang2018posecnn}. 


\subsection{Implementation details}
For each object model we apply the farthest point sampling (FPS) algorithm~\cite{Peng2019pvnet} on the 3D point cloud and select 11 landmarks. The groundtruth 2D landmarks are then obtained by projecting the 3D landmarks with groundtruth camera pose and intrinsics. 
We use ImgAug~\cite{imgaug} for regular data augmentations including rotation, translation, scaling and color jitter before the OBA. We use the Adam optimizer~\cite{Kingma2015adam} and train the model for 250 epochs on LINEMOD and 200 epochs on Occluded-LINEMOD and YCB-Video. We set the landmark verification threshold $\epsilon$ to 1 pixel for all datasets.

\begin{figure}[t]
    \centering
    \includegraphics[width=\linewidth]{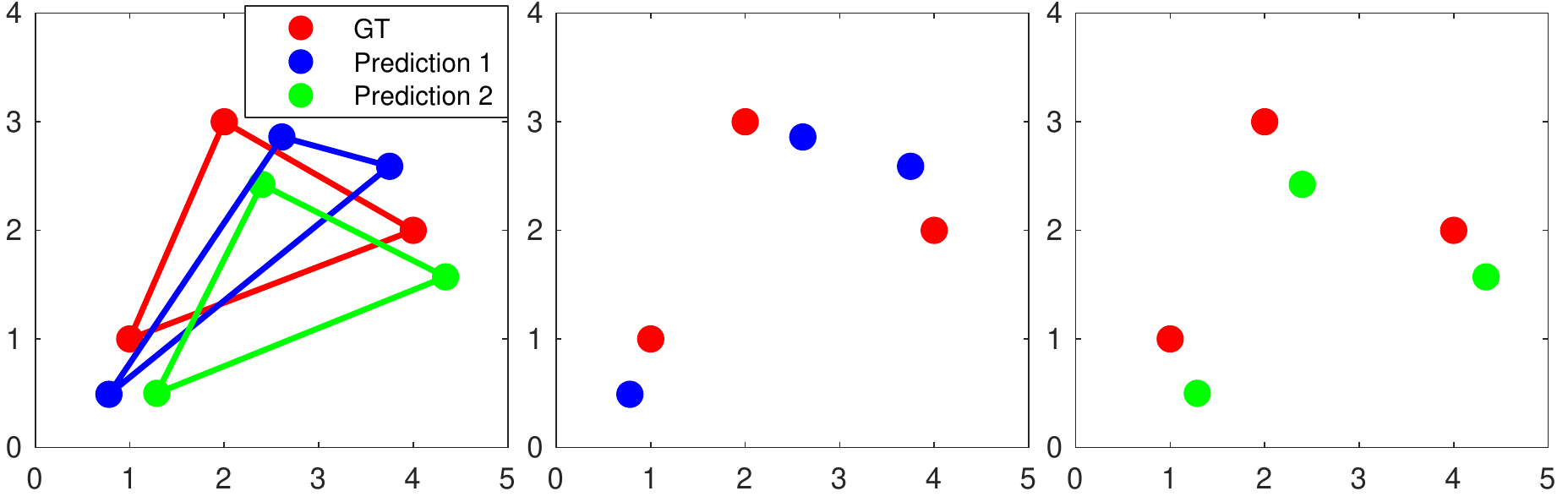}
    \caption{A toy example for the intuition of incoherence measure $c_i$. The mean residual $r_i$ for prediction 1 (blue) and prediction 2 (green) are both 0.608. However, their mean incoherence measure $c_i$ are 0.604 and 0.074, respectively. Although both predictions are identical in terms of accuracy, prediction 2 has much better coherence as the green triangle is much more similar in shape to the groundtruth than the blue one.}
    \label{fig:cohere}
\end{figure}

\subsection{Ablation studies}
\label{sec:ablation}
We conduct various ablation tests to investigate the effect of the proposed OBA and MPS. 

\subsubsection{Model variations}
\label{sec:mvs}

To verify the efficacy of OBA and MPS, we create two Model Variants (MV) of ROPE: 
\begin{enumerate}
    \item (MV1: w/ OBA, w/o MPS) While keeping everything else of the original ROPE unchanged, we remove the low and medium-precision keypoint heads, and train the one-head-model with high-precision groundtruth heatmaps.
    
    \item (MV2: w/o OBA, w/o MPS) On top MV1, we further remove OBA in training. Note that \emph{common data augmentations including rotation, translation, scaling and color jitter, are still kept}. 
\end{enumerate}


Figure~\ref{fig:ablation1} shows the overall ADD(-S) on the Occluded-LINEMOD dataset, as well as qualitative results of all model variants. Without both OBA and MPS, object detection can easily fail and landmark prediction is precarious. We can clearly see that occlusion-robust feature learning enforced by OBA significantly increases the reliability of object detection and landmark prediction. In addition, by comparing MV1 and the original model, it is obvious that MPS boosts the structural consistency of the predicted landmarks, especially in occluded regions. This shows that a holistic representation induced by MPS enhances landmark coherence, strengthening the model's ability to extrapolate to the occluded part of the object.

\begin{figure}[t]
    \centering
    \includegraphics[width=0.95\linewidth]{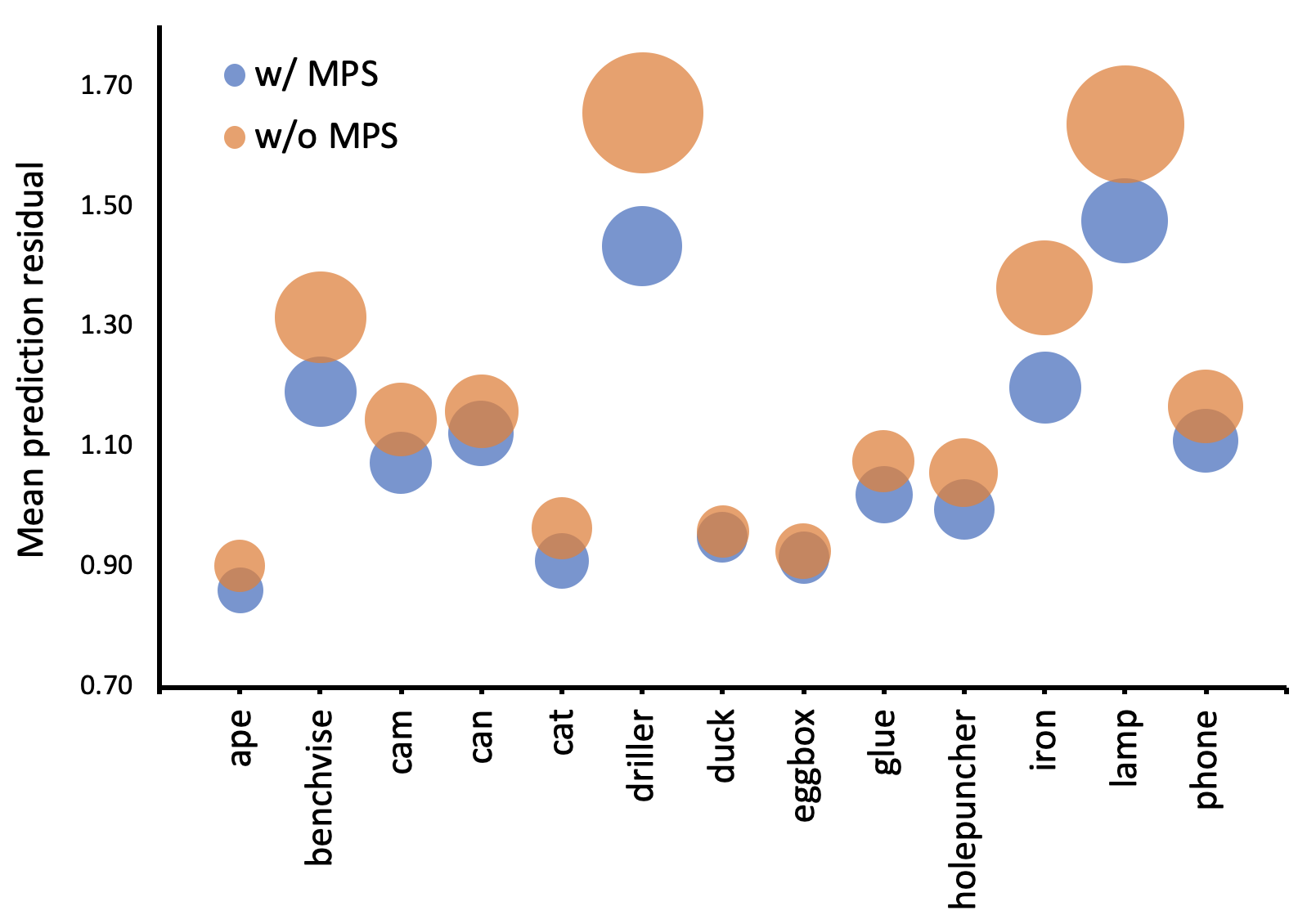}
    \caption{Comparing the results of training with and without MPS on LINEMOD, while keeping all else equal. The vertical location of each bubble represents the mean prediction residual $r_i$ of all landmarks in the testing sets. The size of each bubble indicates the mean incoherence $c_i$.  }
    \label{fig:mps}
\end{figure}

\begin{table*}[t]
    \begin{center}
    \begin{tabular}{l|cccccc|cccc}
   \hline
    \multicolumn{1}{c|}{\multirow{3}{*}{ADD(-S)} }
   &\multicolumn{6}{c|}{Without refinement}
   &\multicolumn{4}{c}{With refinement}
   \\
   \cline{2-11}
  & \multicolumn{1}{c}{PVNet} & \multicolumn{1}{c}{Pix2Pose} & \multicolumn{1}{c}{DPOD}
  & \multicolumn{1}{c}{CDPN}   & \multicolumn{1}{c}{GDR} 
  & \multicolumn{1}{c|}{Ours} & \multicolumn{1}{c}{SSD-6D} & \multicolumn{1}{c}{DPOD+}
  & \multicolumn{1}{c}{HybridPose} & \multicolumn{1}{c}{DeepIM} 
  \\
     & \multicolumn{1}{c}{\cite{Peng2019pvnet}} & \multicolumn{1}{c}{\cite{Park2019pix2pose}} & \multicolumn{1}{c}{\cite{Zakharov2019dpod}}
  & \multicolumn{1}{c}{\cite{Li2019cdpn}} 
  & \multicolumn{1}{c}{\cite{wang2021gdr}} & \multicolumn{1}{c|}{} & \multicolumn{1}{c}{\cite{Kehl2017ssd}} & \multicolumn{1}{c}{\cite{Zakharov2019dpod}}
  & \multicolumn{1}{c}{\cite{song2020hybridpose}} & \multicolumn{1}{c}{\cite{Li2018deepim}} \\
  \hline
ape & 43.62 & 58.10 & 53.28 & 64.38 & - & \textbf{81.52} & 65.00 & \textbf{87.70} & 63.10 & 77.00 \\
benchevise & 99.90 & 91.00 & 95.34 & 97.77 & - & \textbf{100.00} & 80.00 & 98.50 & \textbf{99.90} & 97.50 \\
cam & 86.86 & 60.90 & 90.36 & 91.67 & - & \textbf{96.86} & 78.00 & \textbf{96.10} & 90.40 & 93.50 \\
can & 95.47 & 84.40 & 94.10 & 95.87 & - & \textbf{98.72} & 86.00 & \textbf{99.70} & 98.50 & 96.50 \\
cat & 79.34 & 65.00 & 60.38 & 83.83 & - & \textbf{94.71} & 70.00 & \textbf{94.70} & 89.40 & 82.10 \\
driller & 96.43 & 76.30 & 97.72 & 96.23 & - & \textbf{99.01} & 73.00 & \textbf{98.80} & 98.50 & 95.00 \\
duck & 52.58 & 43.80 & 66.01 & 66.76 & - & \textbf{85.35} & 66.00 & \textbf{86.30} & 65.00 & 77.70 \\
eggbox* & 99.15 & 96.80 & 99.72 & 99.72 & - & \textbf{100.00} & \textbf{100.00} & 99.90 & \textbf{100.00} & 97.10 \\
glue* & 95.66 & 79.40 & 93.83 & \textbf{99.61} & - & 99.42 & \textbf{100.00} & 96.80 & 98.80 & 99.40 \\
holepuncher & 81.92 & 74.80 & 65.83 & 85.82 & - & \textbf{90.39} & 49.00 & 86.90 & \textbf{89.70} & 52.80 \\
iron & 98.88 & 83.40 & 99.80 & 97.85 & - & \textbf{100.00} & 78.00 & \textbf{100.00} & \textbf{100.00} & 98.30 \\
lamp & 99.33 & 82.00 & 88.11 & 97.89 & - & \textbf{99.42} & 73.00 & 96.80 & \textbf{99.50} & 97.50 \\
phone & 92.41 & 45.00 & 74.24 & 90.75 & - & \textbf{97.64} & 79.00 & 94.70 & \textbf{94.90} & 87.70 \\
\hline
average & 86.27 & 72.38 & 82.98 & 89.86 & 93.70 & \textbf{95.61} & 76.69 & \textbf{95.15} & 91.36 & 88.60 \\
\hline
 \end{tabular}
\end{center}
\caption{
Test accuracy on the LINEMOD dataset in terms of the ADD(-S) metric. Objects with a ``*'' sign are considered as symmetric objects and the ADD-S metric is used. The result of SSD-6D is obtained from \cite{tekin2018real}. The result of HybridPose is from its fourth version update in~\cite{song2020hybridposev4}.}
\label{tab:lm}
\end{table*}

\begin{table*}[h]
\begin{center}
\begin{tabular}{l|cccccccc|cc}
   \hline
    \multicolumn{1}{c|}{\multirow{3}{*}{ADD(-S)} }
   &\multicolumn{8}{c|}{Without refinement}
   &\multicolumn{2}{c}{With refinement}
   \\
   \cline{2-11}
   & \multicolumn{1}{c}{HM} & \multicolumn{1}{c}{PVNet} & \multicolumn{1}{c}{Hu}
  & \multicolumn{1}{c}{Pix2Pose} & \multicolumn{1}{c}{DPOD} &
  \multicolumn{1}{c}{Hu2} & \multicolumn{1}{c}{GDR} &
  \multicolumn{1}{c|}{Ours} & \multicolumn{1}{c}{DPOD+}
  & \multicolumn{1}{c}{HybridPose} \\
  & \multicolumn{1}{c}{\cite{Oberweger2018making}} & \multicolumn{1}{c}{\cite{Peng2019pvnet}} & \multicolumn{1}{c}{\cite{Hu2019segmentation}}
  & \multicolumn{1}{c}{\cite{Park2019pix2pose}} & \multicolumn{1}{c}{\cite{Zakharov2019dpod}} &
  \multicolumn{1}{c}{\cite{Hu2020single}} & \multicolumn{1}{c}{\cite{wang2021gdr}} & \multicolumn{1}{c|}{} & \multicolumn{1}{c}{\cite{Zakharov2019dpod}}
  & \multicolumn{1}{c}{\cite{song2020hybridpose}} \\
  \hline
ape & 15.30 & 15.81 & 12.10 & 22.00 & - & 19.20 & \textbf{39.30} & 28.03 & - & \textbf{20.90} \\
can & 44.70 & 63.30 & 39.90 & 44.70 & - & 65.10 & \textbf{79.20} & 75.06 & - & \textbf{75.30} \\
cat & 9.33 & 16.68 & 8.20 & 22.70 & - & 18.90 & 23.50 & \textbf{25.53} & - & \textbf{24.90} \\
driller & 55.40 & 65.65 & 45.20 & 44.70 & - & 69.00 & \textbf{71.30} & 61.86 & - & \textbf{70.20} \\
duck & 19.60 & 25.24 & 17.20 & 15.00 & - & 25.30 & \textbf{44.40} & 19.07 & - & \textbf{27.90} \\
eggbox* & 23.00 & 50.17 & 22.10 & 25.20 & - & 52.00 & \textbf{58.20} & 45.62 & - & \textbf{52.40} \\
glue* & 41.40 & 49.62 & 35.80 & 32.40 & - & 51.40 & 49.30 & \textbf{56.92} & - & \textbf{53.80} \\
holepuncher & 20.40 & 39.67 & 36.00 & 49.50 & - & 45.60 & \textbf{58.70} & 55.54 & - & \textbf{54.20} \\
\hline
average & 28.64 & 40.77 & 27.06 & 32.03 & 32.80 & 43.30 & \textbf{53.00} & 45.95 & 47.30 & \textbf{47.45} \\
\hline
 \end{tabular}
\end{center}
\caption{
Test accuracy on the Occluded-LINEMOD dataset in terms of the ADD(-S) metric. Objects with a ``*'' sign are considered as symmetric objects and the ADD-S metric is used. The result of HybridPose is from its fourth version update in~\cite{song2020hybridposev4}.}
\label{tab:lmo}
\end{table*}

\subsubsection{Accuracy and coherence of landmarks}


To formally analyse the effect of holistic representation learning, we quantify accuracy and structural consistency of landmark predictions and compare them when trained with and without MPS. For accuracy, we define 
\begin{equation}
    \bm r_i = \lVert \bx_i - \bx_i^* \rVert_2
\end{equation}
as the prediction residual of a 2D landmark $\bx_i$. We also define a measure of incoherence 
\begin{equation}
    c_i = \lVert (\bx_i - \bx_i^*)-\bm m \rVert_2
\end{equation}
for a landmark prediction $\bx_i$ where $\bm m = \frac{1}{n}\sum_{i=1}^n (\bx_i - \bx_i^*)$ is the mean error vector for an image. The smaller $c_i$ is, the more coherent a prediction $\bx_i$ is, resulting a more consistent structure of prediction to the groundtruth. An intuitive example is shown in Figure~\ref{fig:cohere}.


As shown in Figure~\ref{fig:mps}, training with MPS effectively lowers the mean residuals. Furthermore, the mean incoherence are also smaller for all objects. This confirms that a more holistic understanding of the object pose can produce more accurate and structurally consistent landmark predictions. 

\begin{table*}[h]
    \begin{center}
    \setlength\tabcolsep{5.5pt}
    \begin{tabular}{l|ccccc||cccc|cc}
   \hline
    \multicolumn{1}{c|}{\multirow{2}{*}{} }
   &\multicolumn{5}{c||}{ADD(-S)}
   &\multicolumn{6}{c}{AUC of ADD(-S)}
   \\
   \cline{2-12}
   \multicolumn{1}{c|}{}
   &\multicolumn{5}{c||}{Without refinement}
   &\multicolumn{4}{c|}{Without refinement}
   &\multicolumn{2}{c}{With refinement}
   \\
   \cline{2-12}
  & \multicolumn{1}{c}{HM} & \multicolumn{1}{c}{Hu} &
  \multicolumn{1}{c}{Hu2} & \multicolumn{1}{c}{GDR} &
  \multicolumn{1}{c||}{Ours}
  & \multicolumn{1}{c}{HM} & \multicolumn{1}{c}{PVNet} & \multicolumn{1}{c}{GDR} & \multicolumn{1}{c|}{Ours}  &
  \multicolumn{1}{c}{DeepIM} & \multicolumn{1}{c}{CosyPose}\\
  & \multicolumn{1}{c}{\cite{Oberweger2018making}} & \multicolumn{1}{c}{\cite{Hu2019segmentation}} & \multicolumn{1}{c}{\cite{Hu2020single}} & \multicolumn{1}{c}{\cite{wang2021gdr}} & \multicolumn{1}{c||}{}
  & \multicolumn{1}{c}{\cite{Oberweger2018making}} & \multicolumn{1}{c}{\cite{Peng2019pvnet}} & \multicolumn{1}{c}{\cite{wang2021gdr}} & \multicolumn{1}{c|}{} & \multicolumn{1}{c}{\cite{li2020deepim}} & \multicolumn{1}{c}{\cite{labbe2020cosypose}}\\
  \hline
master chef can & 31.20 & 33.00 & - & - & \textbf{46.52} & 69.00 & - & - & \textbf{71.17} & \textbf{71.20} & - \\
cracker box & 75.00 & 44.60 & - & - & \textbf{92.63} & 80.20 & - & - & \textbf{89.86} & \textbf{83.60} & - \\
sugar box & 47.20 & 75.60 & - & - & \textbf{99.15} & 76.20 & - & - & \textbf{93.21} & \textbf{94.10} & - \\
tomato soup can & 30.20 & 40.80 & - & - & \textbf{60.90} & 70.00 & - & - & \textbf{82.53} & \textbf{86.10} & - \\
mustard bottle  & 72.50 & 70.60 & - & - & \textbf{100.00} & 84.80 & - & - & \textbf{95.34} & \textbf{91.50} & - \\
tuna fish can  & 4.31 & 18.10 & - & - & \textbf{52.96} & 49.40 & - & - & \textbf{88.01} & \textbf{87.70} & - \\
pudding box  & 48.30 & 12.20 & - & - & \textbf{79.91} & 82.20 & - & - & \textbf{90.5} & \textbf{82.70} & - \\
gelatin box  & 37.20 & \textbf{59.40} & - & - & 58.88 & 81.80 & - & - & \textbf{89.36} & \textbf{91.90} & - \\
potted meat can  & 40.30 & 33.30 & - & - & \textbf{58.62} & 66.20 & - & - & \textbf{74.54} & \textbf{76.20} & - \\
banana  & 6.20 & 16.60 & - & - & \textbf{36.94} & 52.90 & - & - & \textbf{58.77} & \textbf{81.20} & - \\
pitcher base  & 53.80 & 90.00 & - & - & \textbf{99.65} & 69.90 & - & - & \textbf{92.86} & \textbf{90.10} & - \\
bleach cleanser  & 57.20 & 70.90 & - & - & \textbf{75.22} & 73.30 & - & - & \textbf{77.35} & \textbf{81.20} & - \\
bowl*  & \textbf{49.50} & 30.50 & - & - & 45.07 & \textbf{80.30} & - & - & 70.81 & \textbf{81.40} & - \\
mug  & 10.50 & 40.70 & - & - & \textbf{66.04} & 50.50 & - & - & \textbf{89.1} & \textbf{81.40} & - \\
power drill  & 63.00 & 63.50 & - & - & \textbf{94.99} & 78.30 & - & - & \textbf{89.4} & \textbf{85.50} & - \\
wood block*  & 48.20 & 27.70 & - & - & \textbf{55.37} & 65.20 & - & - & \textbf{70.62} & \textbf{81.90} & - \\
scissors  & 0.55 & 17.10 & - & - & \textbf{71.27} & 28.20 & - & - & \textbf{84.82} & \textbf{60.90} & - \\
large marker  & 11.70 & 4.80 & - & - & \textbf{11.73} & 48.20 & - & - & \textbf{53.25} & \textbf{75.60} & - \\
large clamp*  & 12.20 & 25.60 & - & - & \textbf{68.12} & 47.20 & - & - & \textbf{77.1} & \textbf{74.30} & - \\
extra large clamp*  & 17.30 & 8.80 & - & - & \textbf{56.16} & 47.50 & - & - & \textbf{55.19} & \textbf{73.30} & - \\
foam brick*  & 63.80 & 34.70 & - & - & \textbf{68.40} & \textbf{85.60} & - & - & 83.78 & \textbf{81.90} & - \\
  \hline
average & 37.15 & 38.98 & 53.90 & 60.10 & \textbf{66.59} & 66.04 & 73.40 & \textbf{84.40} & 79.88 & 81.90 & \textbf{84.50} \\
\hline
 \end{tabular}
\end{center}
\caption{
Test accuracy on the YCB-Video dataset. Objects with a ``*'' sign are considered as symmetric objects.}
\label{tab:ycbv}
\end{table*}

\subsection{Comparing to SOTA methods}

We report results on the LINEMOD dataset in Table~\ref{tab:lm}. We group methods into two types depending on whether they include a separate refinement step or not. Our method achieves the best average ADD(-S),  as well as the best ADD(-S) on most individual objects. Moreover, our method even outperforms all SOTA methods with refinement, further attesting the power of ROPE.  
The results on the Occluded-LINEMOD dataset are summarised in Table~\ref{tab:lmo}. In the non-refinement group, our method ranked second amongst current SOTA methods overall and best on two individual objects. A sample of qualitative results are provided in Figures~\ref{fig:punch} and \ref{fig:ablation1}.
The results on the YCB-Video dataset are reported in Table~\ref{tab:ycbv}. Without refinement, ROPE has the best performance when evaluated with ADD(-S). 

\subsection{Data efficiency}
The LINEMOD dataset has about 1200 images for each object, which results in approximately 180 images (15\%) for the training set. To supplement such a small training set many methods generate a large amount of synthetic images. For example, 
PVNet~\cite{Peng2019pvnet} renders 20000 images for each object and the same strategy is adopted in~\cite{song2020hybridposev4}. Although we only use a moderate amount of 1312 synthetic images on top of the 180 in training, we test our model's performance in a extremely data-efficient case: only using the $\sim$180 images for training. 

\begin{table} 
    \begin{center}
    \begin{tabular}{l|cc}
   \hline
    \multicolumn{1}{c|}{\multirow{2}{*}{ADD(-S)} }
   &\multicolumn{2}{c}{Training images}
   \\
   \cline{2-3}
  & \multicolumn{1}{c}{$\sim$180} & \multicolumn{1}{c}{$\sim$1500}
  \\
  \hline
ape & 78.57 & \textbf{81.52}  \\
benchvise & 98.93 & \textbf{100.00}   \\
cam & 90.88 & \textbf{96.86}  \\
can & 98.03 & \textbf{98.72}  \\
cat & 92.22 & \textbf{94.71} \\
driller & 98.02 & \textbf{99.01}  \\
duck & 79.06  & \textbf{85.35} \\
eggbox* & 99.72 &  \textbf{100.00}  \\
glue* & 97.68 & \textbf{99.42}  \\
holepuncher & 88.30 & \textbf{90.39} \\
iron & 96.83 & \textbf{100.00}  \\
lamp & 98.85  & \textbf{99.42}  \\
phone & 94.81 &  \textbf{97.64}  \\
\hline
average & 93.22 & \textbf{95.61} \\
\hline
 \end{tabular}
\end{center}
\caption{Comparing performances of ROPE in the extremely data-efficient setting ($\sim$180) and in the original setting ($\sim$1500) on the LINEMOD dataset. Both models are without refinement.}
\label{tab:data_effi}
\end{table}

As shown in Table~\ref{tab:data_effi}, despite having slightly lower ADD(-S) then the baseline, our model achieves an overall accuracy of 93.22\% which is close to the current SOTA method GDR~\cite{wang2021gdr}. This is accomplished with as few as around 180 training images, demonstrating superior data efficiency for our method. 


\section{Conclusion}
We propose ROPE, a framework for robust object pose estimation against occlusions. We show that enforcing occlusion-robust feature learning and encouraging holistic representation learning are the key to achieve occlusion-robustness. Evaluations on three popularly used benchmark datasets, LINEMOD, Occluded-LINEMOD and YCB-Video, show that ROPE either outperforms or is competitive to SOTA methods, without the need of refinement. Our method is also highly data-efficient. 

{\small
\bibliographystyle{ieee_fullname}
\bibliography{egbib}
}

\end{document}